\documentclass[10pt,twocolumn,letterpaper]{article}

\usepackage{cvpr}
\usepackage{times}
\usepackage{epsfig}
\usepackage{graphicx}
\usepackage{amsmath}
\usepackage{amssymb}

\usepackage{booktabs}

\usepackage{helvet} 
\usepackage{courier}  
\usepackage{amssymb,amsmath,amsthm}
\usepackage{mathtools}
\usepackage{mathptmx}

\usepackage{float}
\usepackage{color}
\usepackage{makecell}
\usepackage{multirow}
\usepackage{epstopdf}
\usepackage{colortbl}
\usepackage{pdflscape}
\usepackage{fullpage}
\usepackage{times}
\usepackage{fancyhdr,graphicx,amsmath,amssymb}
\usepackage[ruled,vlined]{algorithm2e}
\include{pythonlisting}

\cvprfinalcopy 

\def\cvprPaperID{8585} 

\ifcvprfinal\fi
\begin{document}

\title{Copy and Paste GAN: Face Hallucination from Shaded Thumbnails}
\author{Yang Zhang$^{1,2,3}$,  Ivor W. Tsang$^{3}$, Yawei Luo$^{4}$, Changhui Hu$^{1,2,5}$, Xiaobo Lu$^{1,2}\footnotemark[1]$,  Xin Yu$^{3,6}$
\\{\small $^{1}$ School of Automation, Southeast University, China }\\{\small $^{2}$ Key Laboratory of Measurement and Control of Complex
Systems of Engineering, Ministry of Education, Southeast University, China }\\{\small $^{3}$ Centre for Artificial Intelligence, University of Technology Sydney, Australia}\\{\small $^{4}$ School of Computer Science and Technology, Huazhong University of Science and Technology, China}\\{\small $^{5}$ School of Automation, Nanjing University of Posts and Telecommunications, China}\\{\small $^{6}$ Australian Centre for Robotic Vision, Australian National University, Australia}
}




\def\XY#1{{\color{red}{\bf [XY:} {\it{#1}}{\bf ]}}}
\def\Yang#1{{\color{blue}{\bf [Yang:} {\it{#1}}{\bf ]}}}
\def\ivor#1{{\color{green}{\bf [Ivor:} {\it{#1}}{\bf ]}}}

\maketitle
\pagestyle{empty}  
\thispagestyle{empty} 

\renewcommand{\thefootnote}{\fnsymbol{footnote}}
\footnotetext[1]{\scriptsize{Corresponding author (xblu2013@126.com).}}
\footnotetext{\scriptsize{This work was done when Yang Zhang (zhangyang201703@126.com) was a visiting student at University of Technology Sydney.}}

\begin{abstract}
Existing face hallucination methods based on convolutional neural networks (CNN) have achieved impressive performance on low-resolution (LR) faces in a normal illumination condition. However, their performance degrades dramatically when LR faces are captured in low or non-uniform illumination conditions. 
This paper proposes a Copy and Paste Generative Adversarial Network (CPGAN) to recover authentic high-resolution (HR) face images while compensating for low and non-uniform illumination. 
To this end, we develop two key components in our CPGAN: internal and external Copy and Paste nets (CPnets). 
Specifically, our internal CPnet exploits facial information residing in the input image to enhance facial details; while our external CPnet leverages an external HR face for illumination compensation.
A new illumination compensation loss is thus developed to capture illumination from the external guided face image effectively.
Furthermore, our method offsets illumination and upsamples facial details alternately in a coarse-to-fine fashion, thus alleviating the correspondence ambiguity between LR inputs and external HR inputs. 
Extensive experiments demonstrate that our method manifests authentic HR face images in a uniform illumination condition and outperforms state-of-the-art methods qualitatively and quantitatively.
\end{abstract}


\section{Introduction}

\begin{figure}[t]
\begin{center}
\includegraphics[width=1\linewidth]{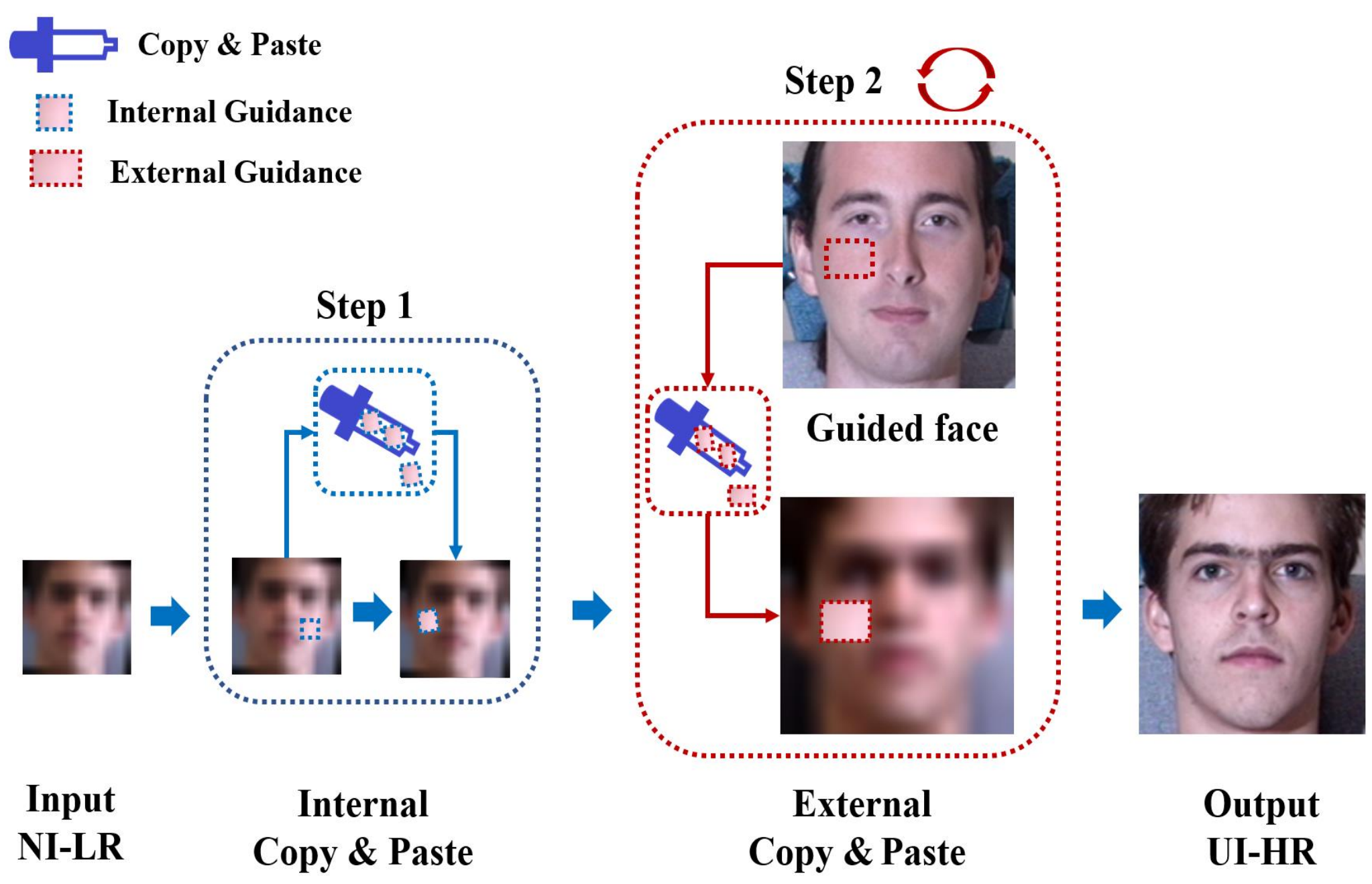}
\vspace{-0.6cm}
\end{center}
   \caption{Motivation of CPGAN.  
   Internal and external CPnets are introduced to mimic  the Clone Stamp Tool.
   Internal CPnet copies good details to paste them on shadow regions. External CPnet further retouches the face using 
   an external guided face from the UI-HR face dataset during the upsampling process to compensate for uneven illumination in the final HR face.}
\vspace{-0.5cm}
\label{fig0}
\end{figure}

\begin{figure*}[htb]
\begin{center}
\includegraphics[width=1\linewidth]{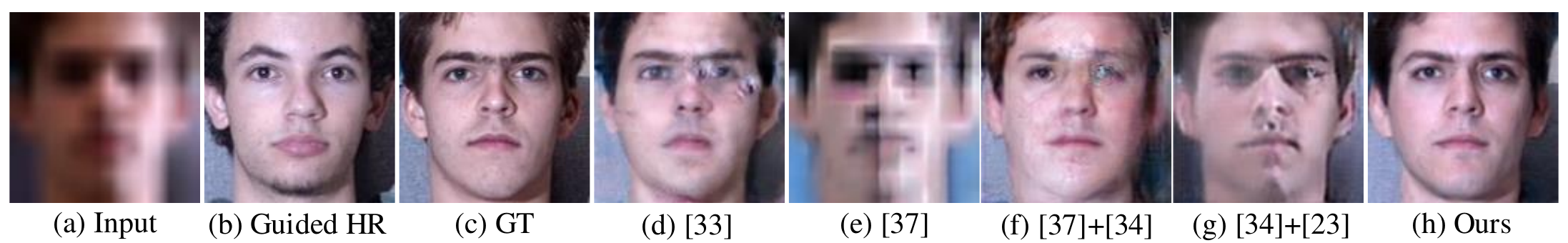}
\end{center}
\vspace{-0.1cm}
 \caption{Face hallucination and illumination normalization results of state-of-the-art methods and our proposed CPGAN. (a) Input NI-LR image ($16\times 16$ pixels); (b) Guided UI-HR image ($128\times 128$ pixels). (c) UI-HR image ($128\times 128$ pixels, {\bf not available in training}). (d) Result of a popular face hallucination method, TDAE~\cite{yu2017face}; (e) Illumination normalization result on (a) by applying~\cite{zhu2017unpaired} after bicubic upsampling; (f) Face hallucination result on (e) by~\cite{yu2017hallucinating}; (g) Face hallucination and illumination normalization result on (a) by~\cite{yu2017hallucinating} and~\cite{shu2018portrait}; (h) Result of CPGAN ($128\times 128$ pixels). \bf Above all, our CPGAN achieves a photo-realistic visual effect when producing authentic UI-HR face images.}
\label{fig1}
\vspace{-0.5cm}
\end{figure*}

Human faces are important information resources since they carry information on identity and emotion changes in daily activities. To acquire such information, high-resolution and high-quality face images are often desirable.  
Due to spacing distance and lighting conditions between cameras and humans, captured faces may be tiny or in poor illumination conditions, thus hindering human perception and computer analysis (Fig.\ref{fig1}(a)). 

Recently, many face hallucination  techniques~\cite{yu2016ultra,zhu2016deep,cao2017attention,chen2018fsrnet,yu2018face,yu2020hallucinating,yu2019semantic,yu2019can,yu2018imagining} have been proposed to visualize tiny face images by assuming uniform illuminations on face thumbnail databases, as seen in Fig.\ref{fig1}(b) and Fig.\ref{fig1}(c). 
However, facial details in shaded thumbnails become obscure in case of low/non-uniform illumination conditions, which leads to failures in hallucination due to inconsistent intensities.
For instance, as Fig.\ref{fig1}(d) shows, the hallucinated result generated by the state-of-the-art face hallucination method~\cite{yu2017face} is semantically and perceptually inconsistent with the ground truth (GT), yielding blurred facial details and non-smooth appearance.

Meanwhile, various methods are proposed to tackle illumination changes on human faces.
The state-of-the-art methods of face inverse lighting~\cite{conde2015efficient,zhou2017label} usually fit the face region to a 3D Morphable Model~\cite{yang2011expression} by facial landmarks and then render illumination.
However, these methods are unsuitable to face thumbnails, because facial landmark cannot be detected accurately in such low-resolution images and erroneous face alignment leads to artifacts in the illumination normalized results. This will increase difficulty to learn the mappings between LR and HR faces. 
Image-to-image translation methods, such as~\cite{isola2017image,zhu2017unpaired}, can be an alternative to transfer illumination styles between faces without detecting facial landmarks.
Due to the variety of illumination conditions, the translation method~\cite{zhu2017unpaired} fails to learn a consistent mapping for face illumination compensation, thus distorting facial structure in the output (Fig.\ref{fig1}(e)).

As seen in Fig.\ref{fig1}(f) and Fig.\ref{fig1}(g), applying either face hallucination followed by illumination normalization or illumination normalization followed by hallucination produces results with severe artifacts. To tackle this problem, our work aims at hallucinating LR inputs under non-uniform low illumination (NI-LR face) while achieving HR in uniform illumination (UI-HR face) in a unified framework.
Towards this goal, we propose a Copy and Paste Generative Adversarial Network (CPGAN). CPGAN is designed to explore internal and external image information to normalize illumination, and to enhance facial details of input NI-LR faces.
We first design an internal Copy and Paste net (internal CPnet) to approximately offset non-uniform illumination features and enhance facial details by searching for similar facial patterns within input LR faces for subsequent upsampling procedure.
Our external CPnet is developed to copy illumination from a HR face template, then pass the illumination information to the input. In this way, our network learns how to compensate for illumination of inputs. To reduce the illumination transferring difficulty, we alternately upsample and transfer the illumination in a coarse-to-fine manner.
Moreover, Spatial Transformer Network (STN)~\cite{jaderberg2015spatial} is adopted to align input NI-LR faces, promoting more effectively feature refinement as well as facilitating illumination compensation.
Furthermore, an illumination compensation loss is proposed to capture the normal illumination pattern and transfer the normal illumination to the inputs. As shown in Fig.~\ref{fig1}(h), the upsampled HR face is not only realistic but also resembles the GT with normal illumination.

The contributions of our work are listed as follows:
\begin{itemize}
\vspace{-2mm}
\item We present the first framework, dubbed CPGAN, to address face hallucination and illumination compensation together, in an end-to-end manner, which is optimized by the conventional face hallucination loss and a new illumination compensation loss. 
\vspace{-2mm}
\item  We introduce an internal CPnet to enhance the facial details and normalize illumination coarsely, aiding subsequent upsampling and illumination compensation. 
\vspace{-2mm}
\item We present an external CPnet for illumination compensation by learning illumination from an external HR face. In this fashion, we are able to learn illumination explicitly rather than requiring a dataset with the same illumination condition.
\vspace{-2mm}
\item A novel data augmentation method, Random Adaptive Instance Normalization (RaIN), is proposed to generate sufficient NI-LR and UI-HR face image pairs. Experiments show that our method achieves photo-realistic UI-HR face images.

\end{itemize}


\section{Related work}

\subsection{Face Hallucination}

Face hallucination methods aim at establishing the intensity relationships between input LR and output HR face images. 
The prior works can be categorized into three mainstreams: holistic-based techniques, part-based methods, and deep learning-based models. 

The basic principle of holistic-based techniques is to represent faces by parameterized models. 
The representative models conduct face hallucination by adopting linear mapping~\cite{wang2005hallucinating}, global appearance~\cite{liu2007face}, subspace learning techniques~\cite{kolouri2015transport}. 
However, they require the input LR image to be pre-aligned and in the canonical pose. 
Then, part-based methods are proposed to extract facial regions and then upsample them. Ma \etal~\cite{ma2010hallucinating} employ position patches from abundant HR images to hallucinate HR face images from input LR ones. 
SIFT flow~\cite{tappen2012bayesian} and facial landmarks~\cite{yang2018hallucinating} are also introduced to locate facial components of input LR images. 

Deep learning is an enabling technique for large datasets, and has been applied to face hallucination successfully. Huang~\etal~\cite{huang2017wavelet} introduce wavelet coefficients prediction into deep convolutional networks to super-resolve LR inputs with multiple upscaling factors. 
Yu and Porikli~\cite{yu2017face} first interweave multiple spatial transformation networks (STNs)~\cite{jaderberg2015spatial} into the upsampling framework to super-resolve unaligned LR faces. Zhu~\etal~\cite{zhu2016deep} develop Cascade bi-network to hallucinate low-frequency and high-frequency parts of input LR faces, respectively. Several recent methods explore facial prior knowledge, such as facial attributes~\cite{yu2018super}, parsing maps~\cite{chen2018fsrnet} and component heatmaps~\cite{yu2018face}, for advanced hallucination results.

However, existing approaches mostly focus on hallucinating tiny face images with normal illumination. Thus, in case of non-uniform illuminations, they usually generate serious blurred outputs.

\subsection{Illumination Compensation}
Face illumination compensation methods are proposed to compensate for the non-uniform illumination of human faces and reconstruct face images in a normal illumination condition.

Recent data driven approaches for illumination compensation are based on the illumination cone~\cite{belhumeur1998set} or the Lambertian Reflectance theory~\cite{basri2003lambertian}. These approaches learn the disentangled representations of facial appearance and mimic various illumination conditions based on the 
modeled illumination parameters.
For instance, Zhou~\etal~\cite{zhou2017label} propose a lighting regression network to simulate various lighting scenes for face images. Shu~\etal~\cite{shu2017neural} propose a GAN framework to decompose face images into physical intrinsic components, geometry, albedo, and illumination base. An alternative solution is the image-to-image translation research~\cite{zhu2017unpaired,Choi_2018_CVPR}. Zhu~\etal~\cite{zhu2017unpaired} propose a cycle consistent network to render a content image to an image with different styles. In this way, the illumination condition of the style image can be transferred to the content image.

However, these methods only compensate for non-uniform illumination without well retaining the accurate facial details, especially when the input face images are impaired or low-resolution. 
Due to the above limitations, simply cascading face hallucination and illumination compensation methods is incompetent to attain high-quality UI-HR faces.

\begin{figure*}[htb]
\begin{center}
\includegraphics[width=1\linewidth]{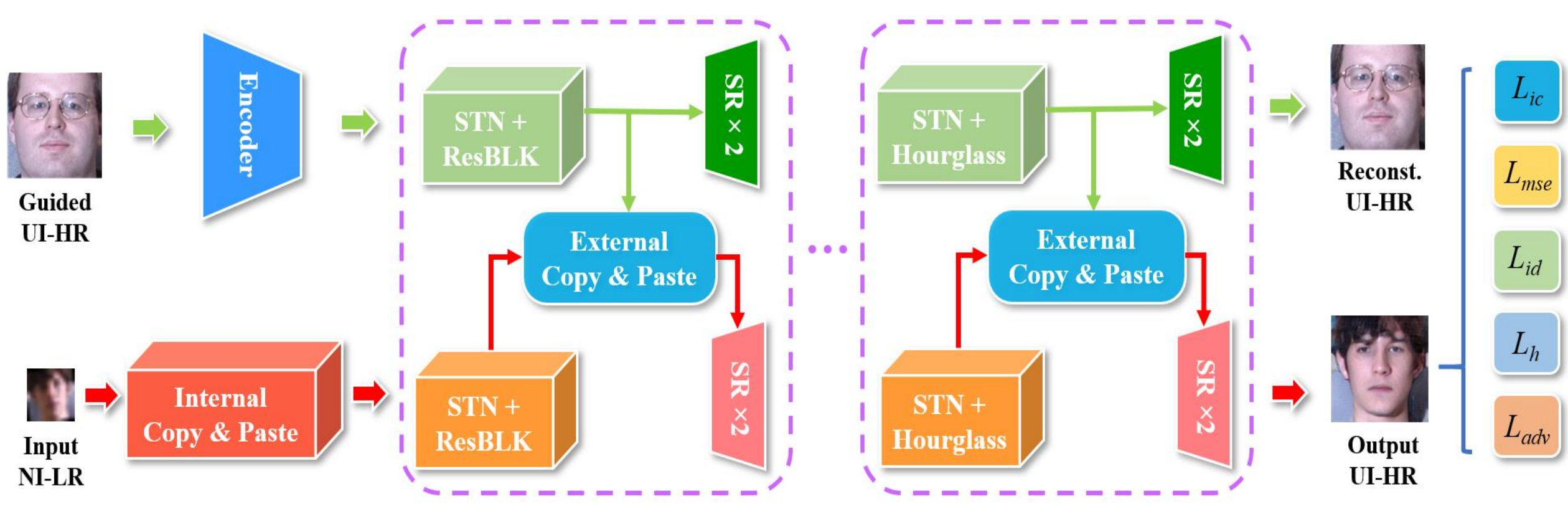}
\end{center}
   \caption{The pipeline of the proposed CPGAN framework. The upper and bottom symmetrical layers in the purple blocks share the same weights.}
\vspace{-0.5cm}
\label{fig3}
\end{figure*}

\section{Hallucination with ``Copy" and ``Paste"}

To reduce the ambiguity of the mapping from NI-LR to UI-HR caused by non-uniform illumination, we present a CPGAN framework that takes a NI-LR face as the input and an external HR face with normal illumination as a guidance to hallucinate a UI-HR one. 
In CPGAN, we develop Copy and Paste net (CPnet) to flexibly ``copy" and ``paste" the  uniform illumination features according to the semantic spatial distribution of the input one, thus compensating for illumination of the input image. A discriminator is adopted to force the generated UI-HR face to lie on the manifold of real face images. The whole pipeline is shown in Fig. \ref{fig3}.

\subsection{Overview of CPGAN}

CPGAN is composed of the following components: internal CPnet, external CPnets, spatial transformer networks (STNs)~\cite{jaderberg2015spatial}, deconvolutional layers, stacked hourglass module~\cite{newell2016stacked} and discriminator network. 
Unlike previous works~\cite{chen2018fsrnet,yu2018face} which only take the LR images as inputs and then super-resolve them with the facial prior knowledge, we incorporate not only input facial information but also an external guided UI-HR face for hallucination. An encoder module is adopted to extract the features of the guided UI-HR image. Note that, our guided face is different from the GT of the NI-LR input. 

As shown in Fig. \ref{fig3}, the input NI-LR image is first passed through the internal CPnet to enhance facial details and normalize illumination coarsely by exploiting the shaded facial information. 
Then the external CPnet resorts to an external guided UI-HR face for further illumination compensation during the upsampling process. 
Because input images may undergo misalignment, such as in-plane rotations, translations and scale changes, we employ STNs to compensate for misalignment~\cite{yu2018face}, as shown in the yellow blocks in Fig. \ref{fig3}. 
Meanwhile, inspired by~\cite{bulat2018super}, we adopt the stacked hourglass network~\cite{newell2016stacked} to estimate vital facial landmark heatmaps for preserving face structure.

\subsubsection{Internal CPnet}

Due to the shading artifacts, the facial details (high-frequency features) in the input NI-LR face image become ambiguous. Therefore, we propose an internal CPnet to enhance the high-frequency features and perform a coarse illumination compensation. 

\begin{figure}[t]
\begin{center}
\includegraphics[width=0.98\linewidth]{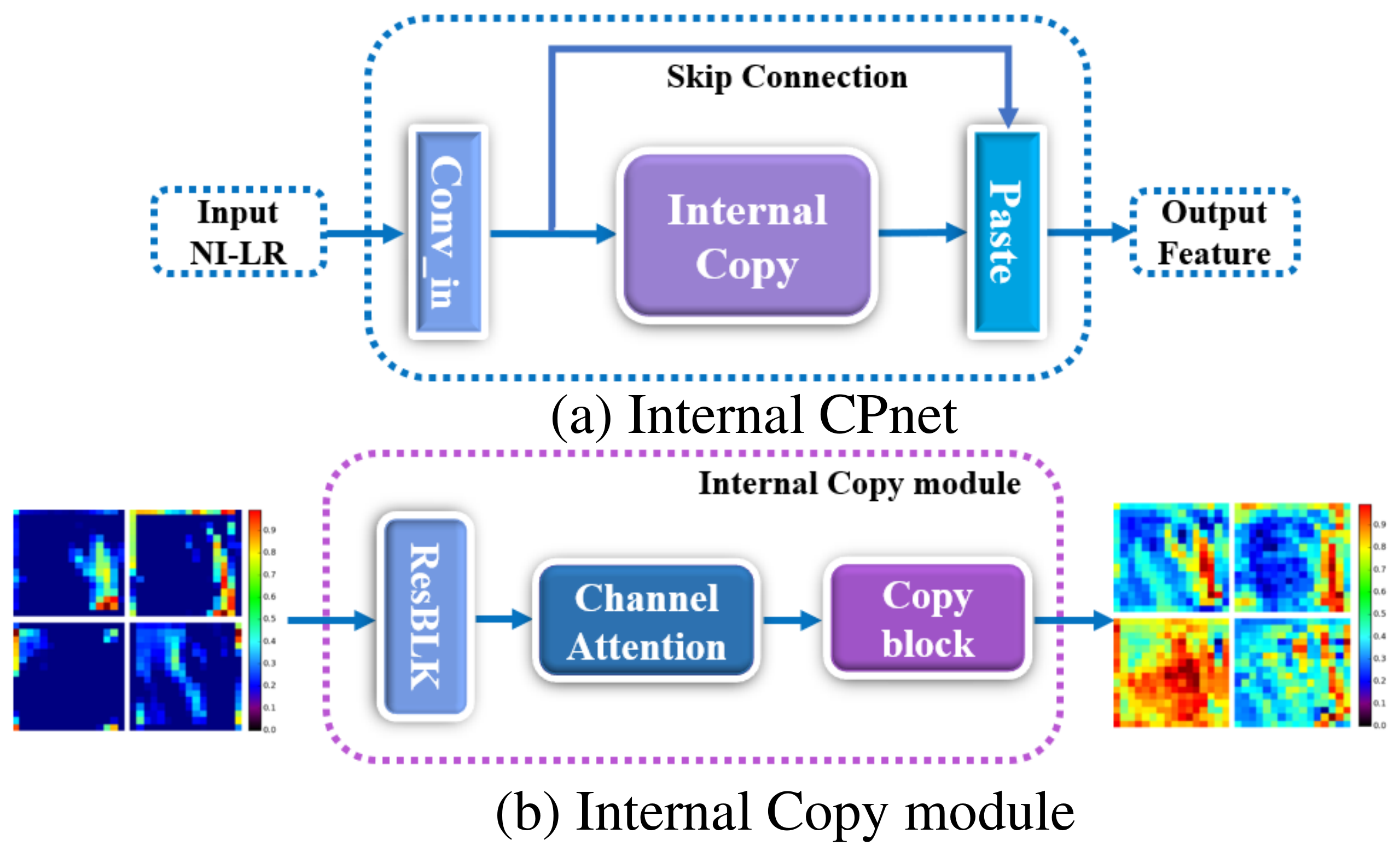}
\end{center}
\vspace{-0.3cm}
 \caption{The architecture of the internal CPnet. Copy block here treats the output features of Channel Attention module as the both input features and guided features. Paste block here represents the additive operation.}
\vspace{-0.5cm}
\label{fig55}
\end{figure}

\begin{figure}[t]
\begin{center}
\includegraphics[width=0.98\linewidth]{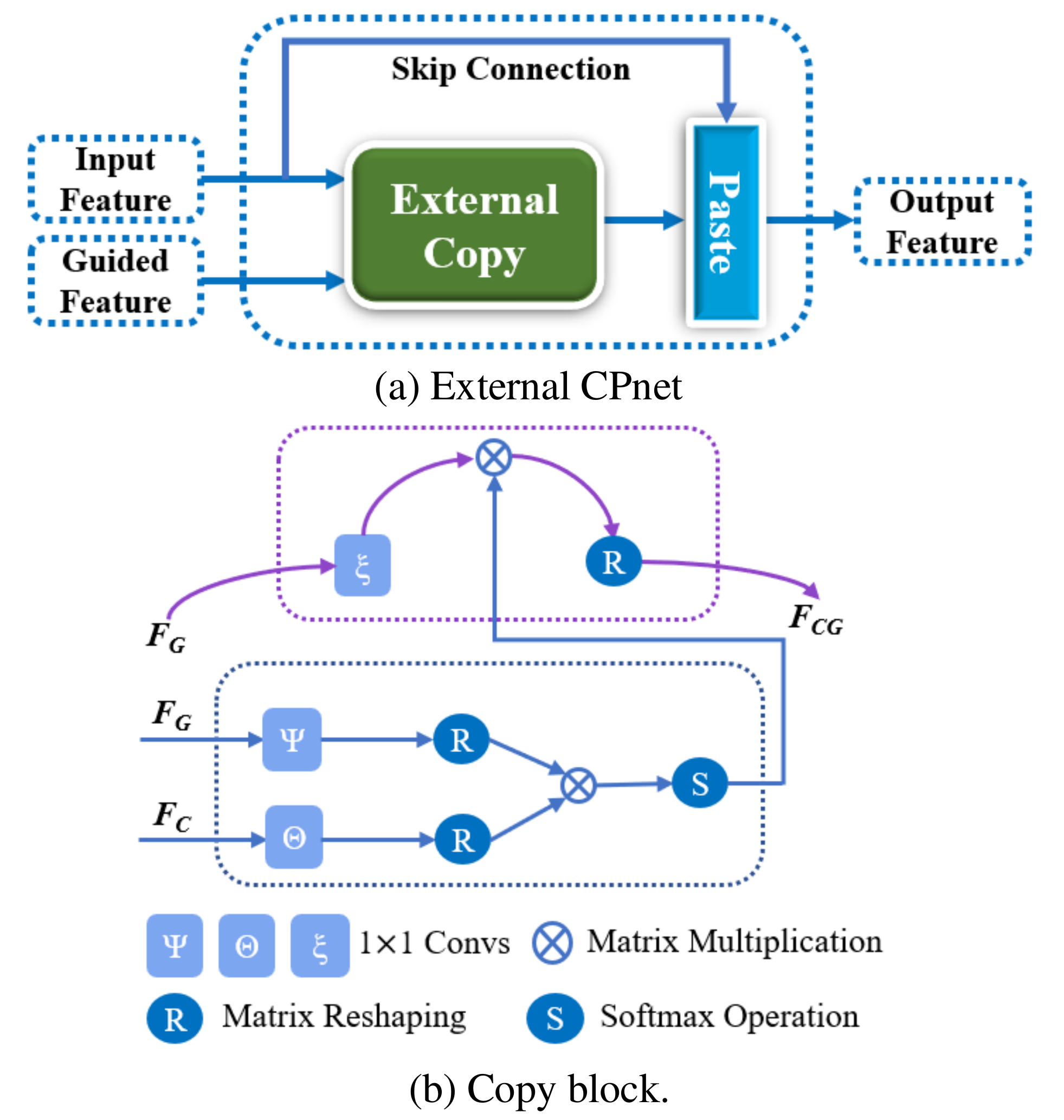}
\end{center}
\vspace{-0.3cm}
\caption{The architecture of the external CPnet. External Copy module here is composed of one Copy block. Paste block represents the additive operation.}
\vspace{-0.5cm}
\label{fig66}
\end{figure}

Fig. \ref{fig55}(b) shows the architecture of our internal CPnet, which consists of an input convolution layer, an Internal Copy module, a Paste block as well as a skip connection. 
Our Internal Copy module adopts the residual block and Channel-Attention (CA) module in~\cite{zhang2018image} to enhance high-frequency features firstly. 
Then, our Copy block (Fig. \ref{fig66}(b)) is introduced to ``copy" the desired internal uniform illumination features for coarsely compensation. 
Note that, Copy block here treats the output features of CA module as the both input features ($F_C$) and guided features ($F_G$) in Fig. \ref{fig66}(b).
Meanwhile, the skip connection in internal CPnet bypasses the LR input features to the Paste block. In this way, the input NI-LR face is initially refined by the internal CPnet. 

To analyze the role of our proposed Internal Copy module, we can exploit the changes of the input and output feature maps. The input feature maps estimate the frequency band of the input NI-LR face, which consists of low-frequency facial components. In this way, they are mainly distributed over the low-frequency band (in blue color). After our Internal Copy module, the output features spread in the direction of high-frequency band (in red color), and literally spans the whole band. 

Thus, we use the name ``Internal Copy module" because its functionality resembles an operation that ``copies" the high-frequency features to the low-frequency parts. Above all, the internal CPnet achieves effective feature enhancement, which benefits subsequent facial detail upsampling and illumination compensation processes.

\vspace{-0.4cm}
\subsubsection{External CPnet}

CPGAN adopts multiple external CPnets and deconvolutional layers to offset the non-uniform illumination and upsample facial details alternately, in a coarse-to-fine fashion. This distinctive design alleviates the ambiguity of correspondences between NI-LR inputs and external UI-HR ones. The network of the external CPnet is shown in Fig. \ref{fig66}(a), and its core components are the Copy and Paste blocks.

Fig. \ref{fig66}(b) performs the ``copy" procedure of the Copy block. The guided features $F_G$ and input features $F_C$ are extracted from the external guided UI-HR image and the input NI-LR image, respectively. First, the guided features $F_G$ and input features $F_C$ are normalized and transformed into two feature space $\theta$ and $\psi$ to calculate their similarity. Then, the ``copied" features $F_{CG}$ can be formulated as a weighted sum of the guided features $F_{G}$ that are similar to the corresponding positions on the input features $F_{C}$. For the $i$ th output response:
\begin{equation}
{F_{CG}^{i}}=\frac{1}{M(F)} \sum_{\forall j} \left\{\exp \left(\boldsymbol{W}_{\theta}^{T} \boldsymbol{\left(\overline{{F_{C}^{i}}}\right)}^{T}\boldsymbol{\overline{{F_{G}^{j}}}} \boldsymbol{W}_{\psi}\right) \boldsymbol{F_{G}^{j}} \boldsymbol{W}_{\zeta}\right\}
\end{equation}
where $M(F)=\sum_{\forall j}\exp\left(\boldsymbol{W}_{\theta}^{T} \boldsymbol{\left(\overline{{F_{C}^{i}}}\right)}^{T}\boldsymbol{\overline{{F_{G}^{j}}}} \boldsymbol{W}_{\psi}\right)$ is the sum of all output responses over all positions. $\overline{F}$ is a transform on $F$ based on the mean-variance channel-wise normalization. Here, the embedding transformations ${W}_{\theta}$, ${W}_{\psi}$ and ${W}_{\zeta}$ are learnt during the training process.

As a result, the Copy block can flexibly integrate the illumination pattern of the guided features into the input features. Based on the Copy and Paste blocks, our proposed external CPnet learns the illumination pattern from the external UI-HR face explicitly. 

\subsection{Loss Function}

To train our CPGAN framework, we propose an illumination compensation loss ($L_{ic}$) together with an intensity similarity loss ($L_{mse}$), an identity similarity loss ($L_{id}$)~\cite{shiri2019identity}, a structure similarity loss ($L_{h}$)~\cite{bulat2018super} and an adversarial loss ($L_{adv}$)~\cite{goodfellow2014generative}. We will detail the illumination loss shortly. For the rest, please refer to the supplementary material. 

The overall loss function $L_{G}$ is a weighted summation of the above terms.
\vspace{-0.3cm}
\begin{equation}
L_{G}=\alpha L_{mse}+\beta L_{id}+\gamma L_{h}+\chi L_{ic}+\kappa L_{adv}
\vspace{-0.3cm}
\label{eq9}
\end{equation}

\noindent\textbf{Illumination Compensation Loss:} CPGAN not only recovers UI-HR face images but also compensates for the non-uniform illumination. Inspired by the style loss in AdaIN~\cite{huang2017arbitrary}, we propose the illumination compensation loss ${L}_{ic}$. The basic idea is to constrain the illumination characteristics of the reconstructed UI-HR face is close to the guided UI-HR one in the latent subspace.
\vspace{-0.3cm}
\begin{equation}
\begin{aligned} 
{L}_{ic} =&\mathbb{E}_{\left(\hat{h}_{i}, g_{i}\right) \sim p(\hat{h}, g)}\{\sum_{j=1}^{L}\left\|\mu\left(\varphi_{j}(\hat{h}_{i})\right)-\mu\left(\varphi_{j}({g}_{i})\right)\right\|_{2} \\ & + \sum_{j=1}^{L}\left\|\sigma\left(\varphi_{j}(\hat{h}_{i})\right)-\sigma\left(\varphi_{j}({g}_{i})\right)\right\|_{2}\}
\end{aligned}
\label{eq90}
\end{equation}
where $g_{i}$ represents the guided UI-HR image, $\hat{h}_{i}$ represents the generated UI-HR image, $p(\hat{h}, g)$ represents their joint distribution. Each $\varphi_{j}(\cdot)$ denotes the output of relu1-1, relu2-1, relu3-1, relu4-1 layer in a pre-trained VGG-19 model~\cite{simonyan2014very}, respectively. Here, $\mu$ and $\sigma$ are the mean and variance for each feature channel.

\section{Data augmentation}

\begin{figure}[tb]
\begin{center}
\includegraphics[width=1\linewidth]{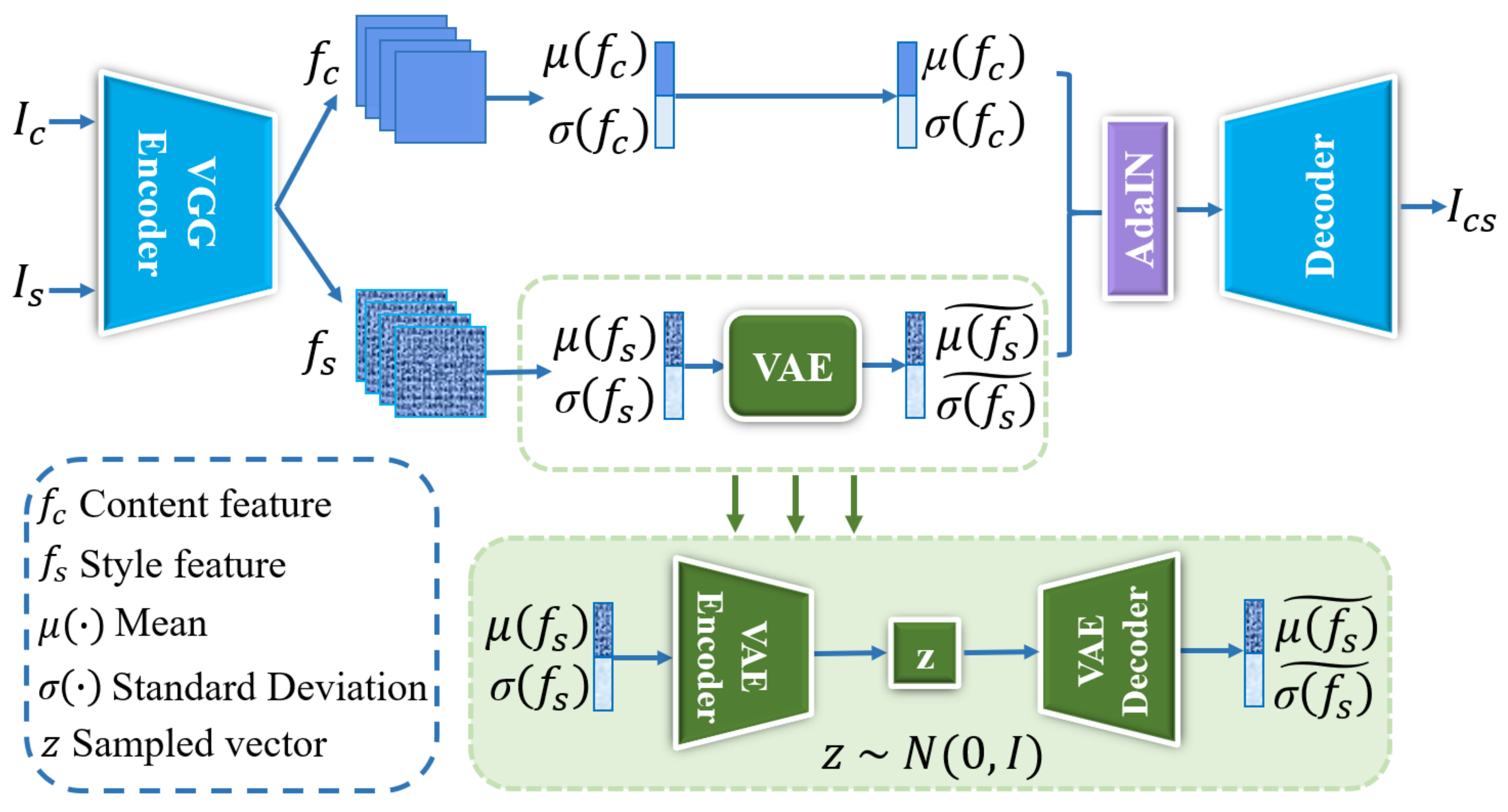}
\vspace{-0.7cm}
\end{center}
   \caption{The training process of RaIN model.}
\vspace{-0.4cm}
\label{fig-adain}
\end{figure}

Training a deep neural network requires a lot of samples to prevent overfitting. 
Since none or limited NI/UI face pairs are available in public face datasets~\cite{gross2010multi,liu2015faceattributes}, we propose a tailor-made Random Adaptive Instance Normalization (RaIN) model to achieve arbitrary illumination style transfer in real-time, generating sufficient samples for data augmentation (Fig. \ref{fig-adain}).

RaIN adopts the encoder-decoder architecture, in which the encoder is fixed to the first few layers (up to relu4-1) of a pre-trained VGG-19~\cite{simonyan2014very}.
The Adaptive Instance Normalization (AdaIN)~\cite{huang2017arbitrary} layer is embedded to align the feature statistics of the UI face image with those of the NI face image. 
Specially, we embed Variational Auto-Encoder (VAE)~\cite{kingma2013auto} before the AdaIN layer. 
In this way, we can efficiently produce an unlimited plausible hypotheses for the feature statistics of the NI face image (limited NI face images are provided in public datasets). 
As a result, sufficient face samples with arbitrary illumination conditions are generated.

Fig. \ref{fig-adain} shows the training process of RaIN model. First, given an input content image $I_{c}$ (UI face) and a style image $I_{s}$ (NI face), the VAE in RaIN encodes the distributions of feature statistics (mean ($\mu$) and variance ($\sigma$)) of the encoded style features $f_{s}$.
In this way, a low-dimensional latent space encodes all  possible variants for style feature statistics. 
Then, based on the intermediate AdaIN layer, the feature statistics ($\mu$ and $\sigma$) of the $f_{c}$ are aligned with a randomly chosen  style feature statistics, forming the transferred feature $t$ via

\vspace{-0.4cm}
\begin{equation}
\begin{aligned}
t=\operatorname{AdaIN}(I_{c}, I_{s})=\widetilde{\sigma(f_{s})}\left(\frac{f_{c}-\mu(f_{c})}{\sigma(f_{c})}\right)+\widetilde{\mu(f_{s})}
\end{aligned}
\label{eq12}
\end{equation}
where $\mu$ and $\sigma$ are computed across each spatial dimension independently for each channel and each sample. 

Then, a randomly initialized decoder $g$ is trained to map $t$ back to the image space, generating the stylized image $I_{cs}$:
\begin{equation}
I_{cs}=g[t]
\label{eq14}
\end{equation}

\begin{figure}[t]
\begin{center}
\includegraphics[width=1\linewidth]{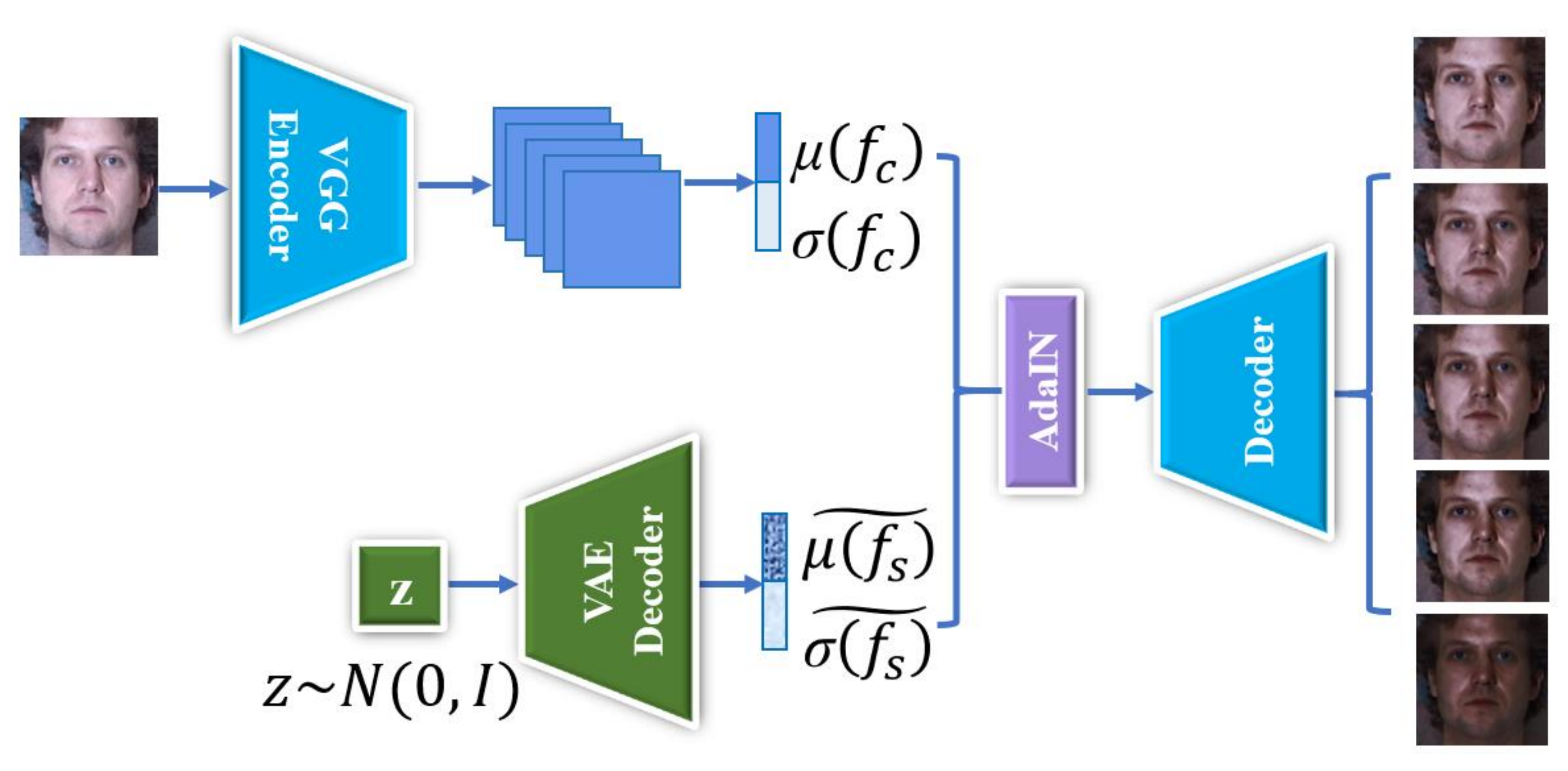}
\vspace{-0.6cm}
\end{center}
   \caption{The generating stage of the RaIN model. A random noise in normal distribution can generate a new style sample.}
\label{fig6}
\vspace{-0.4cm}
\end{figure}

At training stage, RaIN is trained based on the setting of~\cite{huang2017arbitrary}, followed by a fine-tuning procedure on the face dataset to encode the photo-realistic facial details.
To generate the stylized image with a different illumination condition, we can just feed the content image along with a random noise, as shown in Fig. \ref{fig6}. More generated samples are provided in the supplementary material.


\section{Experiments}
In this section, we provide both qualitative and quantitative evaluations on the proposed framework. We conduct the comparisons in the following three scenarios: 

\begin{itemize}
\vspace{-2mm}  
\item FH: Face hallucination methods (SRGAN~\cite{ledig2017photo}, TDAE~\cite{yu2017face}, FHC~\cite{yu2018face});
\vspace{-2mm}
\item IN+FH: Illumination compensation technique (CycleGAN~\cite{zhu2017unpaired}) + Face hallucination methods (SRGAN~\cite{ledig2017photo}, TDAE~\cite{yu2017face}, FHC~\cite{yu2018face}) (use bicubic interpolation procedure to adjust input size);
\vspace{-2mm}
\item FH+IN: Face hallucination methods (SRGAN~\cite{ledig2017photo}, TDAE~\cite{yu2017face}, FHC~\cite{yu2018face}) + Illumination compensation technique (CycleGAN~\cite{zhu2017unpaired}).
\vspace{-2mm}
\end{itemize} 
 
For a fair comparison, we retrain all these methods using our training datasets. 

\begin{figure*}[htb]
\vspace{-0.5cm}
\begin{center}
\includegraphics[width=0.9\linewidth]{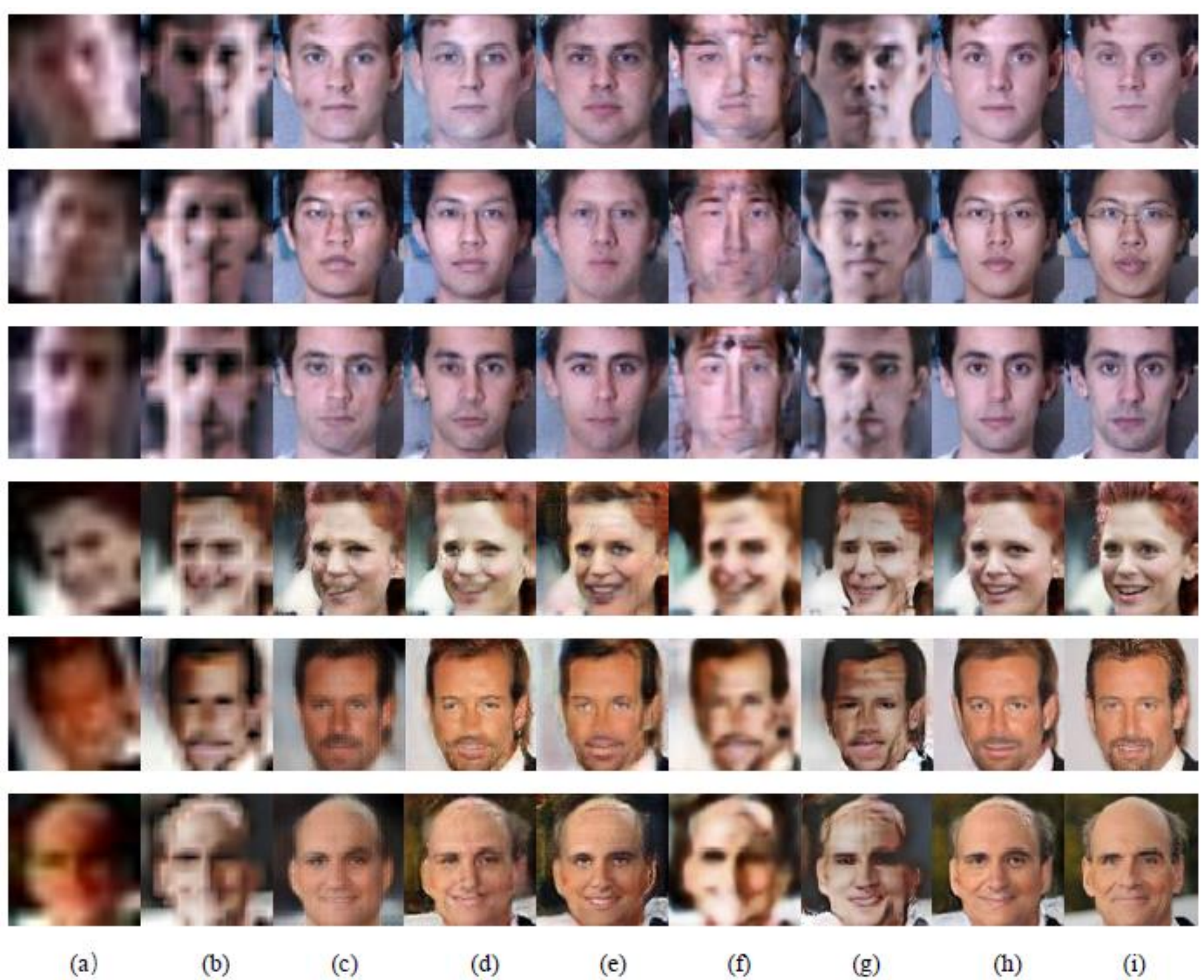}
\vspace{-0.5cm}
\end{center}
   \caption{Comparison with state-of-the-art methods. Columns: (a) Unaligned NI-LR inputs. (b) Bicubic interpolation + CycleGAN~\cite{zhu2017unpaired}. (c) SRGAN~\cite{ledig2017photo}. (d) TDAE~\cite{yu2017face}. (e) FHC~\cite{yu2018face}. (f) CycleGAN~\cite{zhu2017unpaired} + SRGAN~\cite{ledig2017photo}. (g) TDAE~\cite{yu2017face} + CycleGAN~\cite{zhu2017unpaired}. (h) Ours. (i) GT. The first three columns: testing samples from {\bf Multi-PIE} dataset (indoor). The last three columns: testing samples from {\bf CelebA} dataset (in-the-wild).}
\label{fig8}
\end{figure*}

\subsection{Datasets}
CPGAN is trained and tested on the Multi-PIE dataset~\cite{gross2010multi} (indoor) and the CelebFaces Attributes dataset (CelebA)~\cite{liu2015faceattributes} (in the wild).

The Multi-PIE dataset~\cite{gross2010multi} is a large face dataset with 750K+ images for 337 subjects under various poses, illumination and expression conditions. We choose the NI/UI face pairs of the 249 identities under 10 illumination conditions in Session 1 among Multi-PIE dataset, i.e., 249$\times$10 = 2.49K face pairs in total. To enrich the limited illumination conditions, RaIN is adopted to perform data augmentation 10 times on the training set. 

Note that CelebA~\cite{liu2015faceattributes} dataset only provides faces in the wild without NI/UI face pairs. For the training purpose, we opt to generate synthesized NI faces from the UI face images. Similar to~\cite{dong2018style}, Adobe Photoshop Lightroom is adopted to render various illumination conditions. We randomly select 18K NI faces from CelebA dataset to perform rendering. Then, 18K NI/UI face pairs are generated. 

For the GT images, we crop the aligned UI-HR faces and then resize them to $128\times 128$ pixels. For the NI-LR faces, we resize the UI-HR face images to $128\times 128$ pixels and then apply a sequence of manual transformations, including rotations, translations, scaling and downsampling processes to obtain images of $16\times 16$ pixels. We choose 80 percent of the face pairs for training and 20 percent of the face pairs for testing. Specially, during the training process of CPGAN, we randomly select UI-HR images in the training set to serve as the external guided UI-HR images. We will release our synthesized NI/UI face pairs for academic and commercial applications.

\begin{table*}[htb]
\caption{Average PSNR [dB] and SSIM values of compared methods on the testing datasets.}
\begin{center}
\begin{footnotesize}
\begin{tabular}{@{}ccccccccccccc@{}}
\toprule
\multirow{3}{*}{Method} & \multicolumn{2}{c}{Multi-PIE} & \multicolumn{2}{c}{CelebA} & \multicolumn{2}{c}{Multi-PIE} & \multicolumn{2}{c}{CelebA} & \multicolumn{2}{c}{Multi-PIE} & \multicolumn{2}{c}{CelebA} \\ \cmidrule(l){2-13} 
                        & \multicolumn{4}{c}{FH}                                     & \multicolumn{4}{c}{IN+FH}                                  & \multicolumn{4}{c}{FH+IN}                                  \\
                        & PSNR           & SSIM         & PSNR         & SSIM        & PSNR           & SSIM         & PSNR         & SSIM        & PSNR           & SSIM         & PSNR         & SSIM        \\ \bottomrule
Bicubic                 & 12.838         & 0.385        & 12.79        & 0.480       & 13.960         & 0.386        & 13.018       & 0.494       & 13.315         & 0.399        & 12.945       & 0.352       \\
SRGAN                   & 16.769         & 0.396        & 17.951       & 0.536       & 14.252         & 0.408        & 16.506       & 0.488       & 15.044         & 0.432        & 15.315       & 0.398       \\
TDAE                    & 19.342         & 0.411        & 19.854       & 0.530       & 15.449         & 0.445        & 18.031       & 0.540       & 15.319         & 0.427        & 15.727       & 0.403       \\
FHC                     & 20.680         & 0.467        & 21.130       & 0.552       & 17.554         & 0.512        & 19.508       & 0.499       & 18.263         & 0.545        & 16.527       & 0.426       \\\cmidrule(r){1-13}
CPGAN                    &\textbf{24.639}         &\textbf{0.778}        & \textbf{23.972}       &\textbf{0.723}       &\textbf{24.639}         & \textbf{0.778}        &\textbf{23.972}       &\textbf{0.723}       &\textbf{24.639}         &\textbf{0.778}        &\textbf{23.972}       & \textbf{0.723}       \\ \bottomrule
\end{tabular}
\end{footnotesize}
\end{center}
\vspace{-0.7cm}
\label{table1}
\end{table*}

\begin{figure*}[t]
\begin{center}
\includegraphics[width=0.95\linewidth]{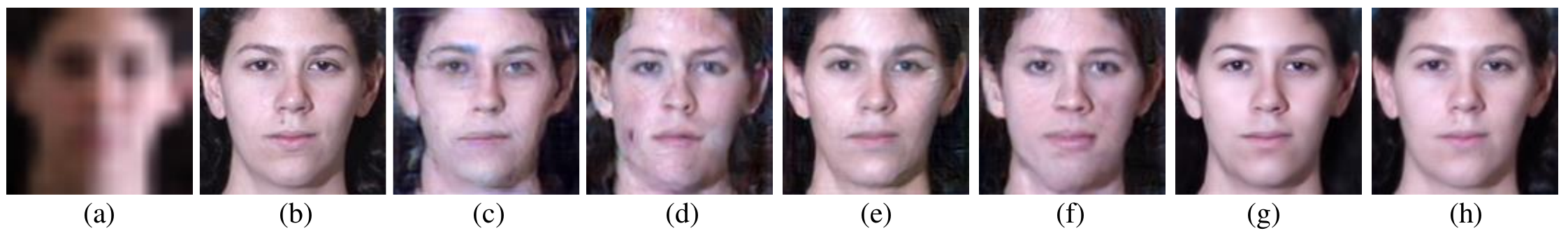}
\vspace{-0.5cm}
\end{center}
\caption{Ablation Analysis. (a)$16\times 16$ NI-LR image. (b)$128\times 128$ GT image. (c) Result without using the internal CPnet, using a simple input convolution layer instead. (d) Result without using the external CPnets, connecting input and output features directly. (e) Result without using the identity similarity loss ($L_{id}$). (f) Result without using structure similarity loss ($L_{h}$). (g) Result without using adversarial loss ($L_{adv}$). (h) CPGAN result.}
\label{fig9}
\vspace{-0.2cm}
\end{figure*}

\begin{figure}[t]
\vspace{-0.2cm}
\begin{center}
\includegraphics[width=0.98\linewidth]{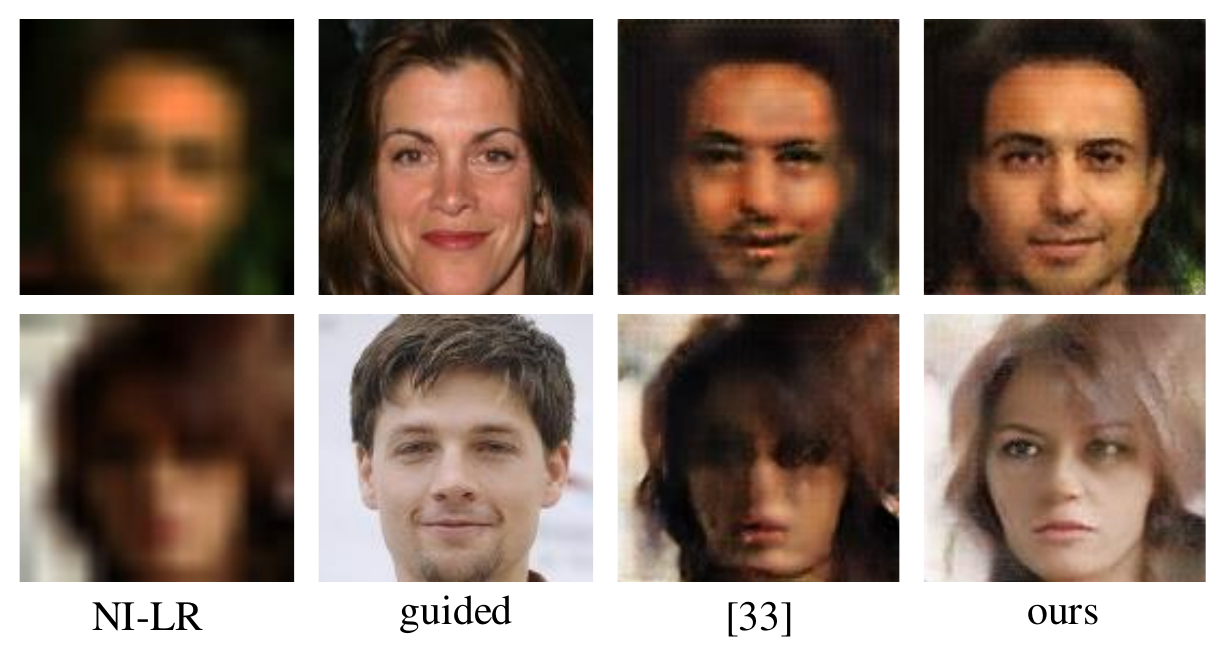}
\vspace{-0.5cm}
\end{center}
\caption{Results on real NI-LR face images.}
\vspace{-0.6cm}
\label{fig10}
\end{figure}

\subsection{Qualitative Comparison with the SoA}
A qualitative comparison with the benchmark methods is presented in Fig.\ref{fig8}, which justifies the supeior performance of CPGAN over the competing methods. Obviously, the image obtained by CPGAN is more authentic, identity-preserving, and having richer facial details.

As illustrated in Fig.\ref{fig8}(b), the combination
of bicubic interpolation and CycleGAN cannot produce authentic face images. Due to the incapability of bicubic interpolation to generate necessary high-frequency facial details and the lack of GT, CycleGAN achieves image-to-image translation with deformed face shapes.

SRGAN~\cite{ledig2017photo} provides an upscaling factor of $8\times$, but it is only trained on general patches. Thus, we retrain SRGAN on face images as well. As shown in Fig.\ref{fig8}(c), the results of SRGAN are still blurred.

TDAE~\cite{yu2017face} super-resolves very low resolution and unaligned face images. It employs deconvolutional layers to upsample LR faces and a discriminative network to promote the generation of sharper results. However, it does not take illumination into account, thus, the final outputs suffer from severe artifacts (Fig.\ref{fig8}(d)).

Yu and Porikli~\cite{yu2018face} exploit the FHC to hallucinate unaligned LR face images. As visualized in Fig.\ref{fig8}(e), FHC fails to construct realistic facial details due to inaccurate facial prior prediction caused by shading artifacts. 

We provide experiments on the IN+FH and FH+IN combination as well for completeness. However, due to the deteriorated super-resolved facial patterns caused by the misleading mappings, their generated results are contaminated with ghosting artifacts, as shown in Fig.\ref{fig8}(f), and Fig.\ref{fig8}(g). 

In contrast, our method is able to reconstruct authentic facial details as shown in Fig.\ref{fig8}(h). Albeit there
exists non-uniform illumination in the input NI-LR faces, our method still produces visually pleasing UI-HR faces
which are close to the GT faces without suffering blurs. For instance, we hallucinate the shading artifacts covered facial parts preciously, such as the jaw and mouth, as illustrated in the third and fourth rows in Fig.\ref{fig8}(h).

\subsection{Quantitative Comparison with the SoA}
The above qualitative performances are verified by the quantitative evaluations. We report the average peak signal-to-noise ratio (PSNR) and structural similarity (SSIM) over three combinations (FH, IN+FH, and FH+IN) on the Multi-PIE and CelebA testing sets.

From Table \ref{table1}, the proposed CPGAN performs unarguably better than the rest in both indoor and in the wild datasets. For instance, in FH scenario, CPGAN outperforms the second best technique with a large margin of approximate 4 dB in PSNR on Multi-PIE dataset.
Thanks to the deliberate design of the internal and external CPnets, the realistic facial details are well recovered from the shading artifacts. 

\subsection{Performance on Real NI-LR faces}
Our method can also effectively hallucinate the real NI-LR faces beyond the synthetically generated face pairs. To demonstrate this, we randomly choose face images with non-uniform illumination from the CelebA dataset. As shown in Fig.\ref{fig10}, our CPGAN can hallucinate such randomly chosen shaded thumbnails, demonstrating its robustness in various circumstances.

\subsection{Ablation Analysis}
\vspace{-2mm}
\textbf{Effectiveness of internal CPnet:} As shown in Fig.\ref{fig9}(c), it can be seen that the result without internal CPnet suffers from severe distortion and blur artifacts. This is because the distinctive designed internal CPnet enhances the facial details in the input NI-LR image, and it aids subsequent upsampling and illumination compensation. A quantitative study is reported in Table \ref{tablea}.

\textbf{Effectiveness of external CPnet:} In our method, external CPNet introduces a guided UI-HR image for illumination compensation. We demonstrate the effectiveness in Fig.\ref{fig9}(d) and Table \ref{tablea}. Without external CPnet, the reconstructed face deviates from the GT appearance seriously, especially for the input part affected by shading artifacts.
It implies that external CPnet learns the illumination pattern from the external UI-HR face explicitly.

\begin{table}[t]
\vspace{-0.4cm}
\caption{Ablation study of CPnet}
\centering
\begin{footnotesize}
\begin{tabular}{@{}cccccc@{}}
\bottomrule
\multirow{2}{*}{\begin{tabular}[c]{@{}c@{}}w/o CPnet\end{tabular}} & \multicolumn{2}{c}{Multi-PIE}    & \multicolumn{2}{c}{CelebA}       \\ \cline{2-5} 
                                                                        & PSNR            & SSIM           & PSNR            & SSIM           \\\bottomrule 
internal CPnet                                                          & 22.164          & 0.693          & 21.441          & 0.578          \\
external CPnet                                                          & 21.925          & 0.680          & 21.032          & 0.536          \\ \bottomrule
CPGAN                                                                   & \textbf{24.639} & \textbf{0.778} & \textbf{23.972} & \textbf{0.723} \\ \bottomrule
\end{tabular}
\end{footnotesize}
\vspace{-0.4cm}
\label{tablea}
\end{table}

\textbf{Loss Function Specifications:} Fig.\ref{fig9} also illustrates the perceptual performance of different training loss variants. Without the identity similarity loss ($L_{id}$) or structure similarity loss ($L_{h}$), the hallucinated facial contours are blurred (Fig.\ref{fig9}(e) and Fig.\ref{fig9}(f)). The adversarial loss ($L_{adv}$) makes the hallucinated face sharper and more realistic, as shown in Fig.\ref{fig9}(h). The effect of our illumination compensation loss ($L_{ic}$) is provided in the supplementary material.



\vspace{-3mm}
\section{Conclusion}
\vspace{-2mm}
This paper presents our CPGAN framework to jointly hallucinate the NI-LR face images and compensate for the non-uniform illumination seamlessly. With the internal and external CPNets, our method enables coarse-to-fine feature refinement based on the semantic spatial distribution of the facial features. In this spirit, we offset shading artifacts and upsample facial details alternately. Meanwhile, the RaIN model mimics sufficient face pairs under diverse illumination conditions to facilitate practical face hallucination. Experimental results validate the effectiveness of CPGAN, which yields photo-realistic visual quality and promising quantitative performance compared with the state-of-the-art.


{\small

\bibliographystyle{ieee_fullname}
}


\end{document}